\def\year{2021}\relax
\documentclass[letterpaper]{article} 
\usepackage{aaai21}  
\usepackage{times}  
\usepackage{helvet} 
\usepackage{courier}  
\usepackage[hyphens]{url}  
\usepackage{graphicx} 
\usepackage{natbib}  
\usepackage{caption} 
\usepackage{times}
\usepackage{soul}
\usepackage{url}
\usepackage[hidelinks]{hyperref}
\usepackage[utf8]{inputenc}
\usepackage{graphicx}
\usepackage{amsmath}
\usepackage{amsfonts}
\usepackage{amsthm}
\usepackage{booktabs}
\usepackage{tablefootnote}
\usepackage{threeparttable}
\usepackage{multirow}
\usepackage{algorithm}
\usepackage{algorithmic}
\usepackage{wrapfig}
\usepackage[switch]{lineno}  %

\newtheorem{lemma}{Lemma}

\newtheorem{theorem}{Theorem}
\newtheorem{proposition}{Proposition}
\usepackage{xcolor}

\title{MimicNorm: Weight Mean and Last BN Layer Mimic the Dynamic of Batch Normalization}

\author {
        Wen~Fei,\textsuperscript{\rm 1}
        Wenrui~Dai, \textsuperscript{\rm 2}
        Chenglin~Li,\textsuperscript{\rm 1}
        Junni~Zou, \textsuperscript{\rm 2}
        Hongkai~Xiong \textsuperscript{\rm 1, \rm 2} \\
}
\affiliations {
    \textsuperscript{\rm 1} the Department of Electronic Engineering, Shanghai Jiao Tong University, Shanghai 200240, China\\
    \textsuperscript{\rm 2} the Department of Computer Science and Engineering, Shanghai Jiao Tong University, Shanghai 200240, China \\
    fw.key@sjtu.edu.cn, daiwenrui@sjtu.edu.cn, 	lcl1985@sjtu.edu.cn, zou-jn@cs.sjtu.edu.cn,
    xionghongkai@sjtu.edu.cn
}

\begin{document}
\maketitle
\begin{abstract}
  Substantial experiments have validated the success of Batch Normalization (BN) Layer in benefiting convergence and generalization. However, BN requires extra memory and float-point calculation. Moreover, BN would be inaccurate on micro-batch, as it depends on batch statistics.
  In this paper, we address these problems by simplifying BN regularization while keeping two fundamental impacts of BN layers, \emph{i.e.}, data decorrelation and adaptive learning rate. We propose a novel normalization method, named \emph{MimicNorm}, to improve the convergence and efficiency in network training. MimicNorm consists of only two light operations, including modified weight mean operations (subtract mean values from weight parameter tensor) and one BN layer before loss function (last BN layer). We leverage the neural tangent kernel (NTK) theory to prove that our weight mean operation whitens activations and transits network into the chaotic regime like BN layer, and consequently, leads to an enhanced convergence. The last BN layer provides auto-tuned learning rates and also improves accuracy. Experimental results show that MimicNorm achieves similar accuracy for various network structures, including ResNets and lightweight networks like ShuffleNet, with a reduction of about 20\% memory consumption. The code is publicly available at https://github.com/Kid-key/MimicNorm.    
  
\end{abstract}

\section{Introduction}

Deep learning applications have led to revolutionary consequences. In addition to the development of hardware, competent algorithms for training complex deep networks play a crucial role in practical scenarios. Batch Normalization (BatchNorm or BN) has emerged as one of the most successful techniques to improve the accuracy and speed up training. BN layers are adopted in most state-of-the-art neural networks to stabilize the training process, especially for deep networks, \emph{e.g.}, ResNets~\cite{he2016deep}. Inspired by BN regulation, a variety of normalization techniques have been developed for natural language processing and unsupervised tasks, including Instance Normalization~\cite{ulyanov2016instance}, Layer Normalization~\cite{ba2016layer}, and Group Normalization~\cite{wu2018group}.

Despite widespread adoption of BN layers, little consensus has been reached on understanding them. 
\citet{ioffe2015batch} normalize activations in intermediate layers of deep neural networks (DNNs) to reduce the internal covariate shift (ICS) and constrain the distribution on each layer. 
However, \citet{santurkar2018does} argue that the effectiveness of BN has little relation to ICS and identify the key impact of BN as the smaller Lipschitz constant and smoother landscape.
\citet{arora2018theoretical} consider the auto rate-tuning mechanism based on the scale-invariant property of BN networks.
The BN layers automatically decrease the learning rates and enhance the convergence speed to a first-order stationary point.
\citet{luo2018towards} demonstrate that BN equivalently imposes Population Normalization (PN) and gamma decay. BN layers also show a tangled impact with weight decay regularization \cite{zhang2018three}.

In this paper, we focus on the mechanism of BN as the guidance for efficient network design. BN layers require buffers for both forward and backward pass that account for 20\% total memory consumption of the networks. Specifically, we aim at addressing the problems listed as below.

\begin{enumerate}
    \item \textbf{Fundamental impacts of BN layers.} BN network normalizes activations into zero mean and unit variance in a  layer-by-layer fashion in the forward pass, and adjusts the gradient propagation in the backward process. It is of necessity to disentangle and identify the fundamental impacts by BN layers on network training.
    \item{\textbf{Efficient  alternative to BN layers.} Simplified and pure regularization can be developed as an efficient alternative to BN layers by maintaining the fundamental impacts and reducing the memory consumption. }
\end{enumerate}



\begin{figure*}[t]
  \centering
  \includegraphics[width=0.6\textwidth]{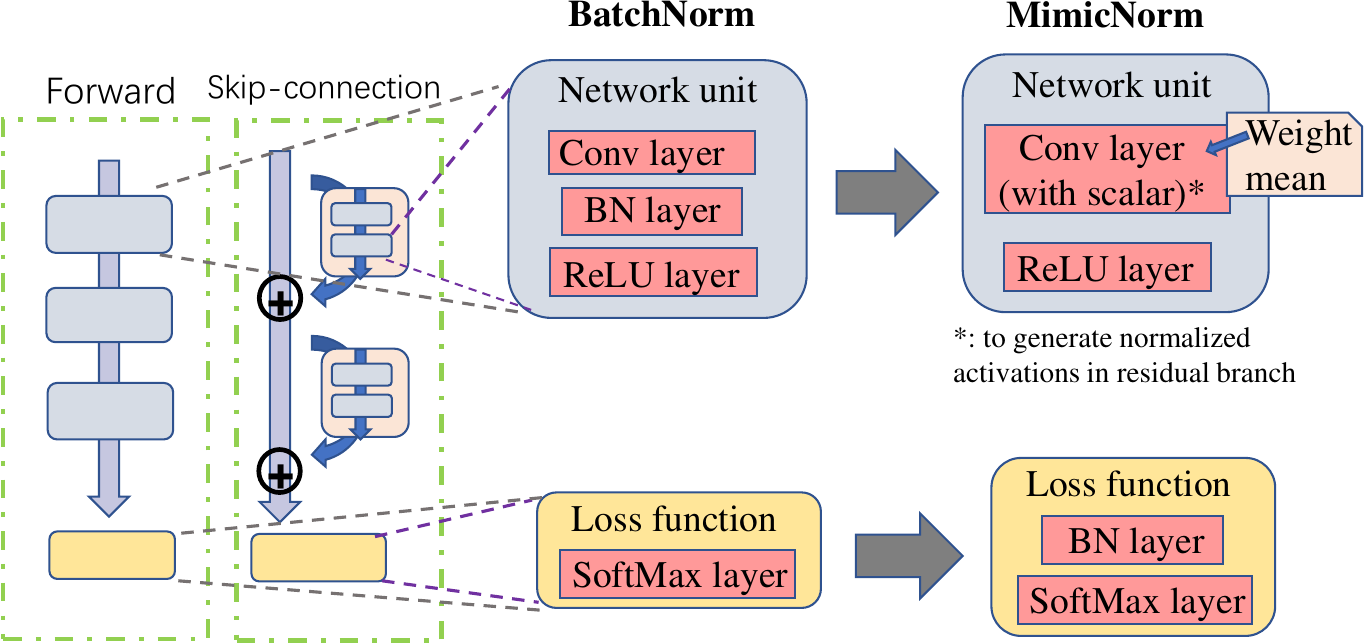}
  \caption{Illustrative diagram of \emph{MimicNorm}. Modern neural networks are composed of stacked basic units and a loss function, usually CrossEntropy for classification. To remove BN layers, we introduce weight mean for each convolutional layers and modify initialization to generate stable intermediate activations. Besides, we apply a learnable scalar multiplier for skip-connection to normalize residual. And one last BN layer is inserted before the loss function to adjust the learning rate.}
  \label{fig.mimic}
\end{figure*} 

Neural tangent kernel (NTK) leverages the kernel method to analyze infinite-width networks~\cite{jacot2018ntk}. NTK is determined by the correlations of gradients to control the network convergence. 
In this paper, we investigate the impact of BN layers on correlation propagation. We demonstrate in theory that both weight mean and BN layers lead to chaotic NTK~\cite{jacot2019freeze}, and consequently, achieve linear convergence rates.
Furthermore, we empirically find that one BN layer before loss function (last BN layer in the rest of this paper) avoids uncontrolled variance and enables large learning rates. The contributions of this paper are summarized as below.

\begin{enumerate}
    \item We demonstrate that, for deep infinite-width networks with ReLU activation function, weight mean operations and BN layers have same NTK expressions and enable network training with equivalent linear convergence rates, as elaborated in Section~\ref{sec.wm}. 
    \item We propose an efficient alternative to BN networks, namely \emph{MimicNorm}, that consists of the weight mean and last BN layer to enable automatic tuning of learning rates, as presented in Figure~\ref{fig.mimic} and Section~\ref{sec.mimicnorm}. 
\end{enumerate}

Experimental results show that MimicNorm achieves the same accuracy on various BN networks with a reduction of memory consumption. For skip-connection structures (\emph{e.g.}, ResNet50 and ResNet110) and lightweight networks (\emph{e.g.}, ShufflNetV2), our method reduces the memory consumption by 20\% in comparison to BN networks.

\section{Background}
We begin with an overview of recent development of neural tangent kernel and the related concepts to describe the training process of DNNs~\cite{jacot2018ntk,jacot2019freeze,lee2019linearmodels,xiao2019disentangling}. For brevity, we keep our discussion within deep Fully-Connected Neural Networks (FCNNs). Similar but intricate analysis of convolutional networks can be found in \cite{arora2019cntk,bietti2019inductive,yang2019scaling}. A general fully-connected network can be represented by the mapping from data space $\mathbb{R}^{n_0}$ to output space $\mathbb{R}^{n_L}$ consisting of $L$ cascade hidden layers.

\begin{align}
    &h^{1}=\frac{\sigma_w}{\sqrt{2n_0}}\mathbf{W}^{0}x+\sigma_b b^0 \ ,  \nonumber\\
h^{l+1}&=\frac{\sigma_w}{\sqrt{n_{l}}}\mathbf{W}^{l}\phi(h^{l})+\sigma_b b^l \quad 1 \le l< L,
\end{align}
where $\phi$ is the nonlinear element-wise function, and we take ReLU activation function $\phi(x)=\max(x,0)$ in this paper. Each layer has weight matrix $\mathbf{W}^l$ and bias vector $b^l$ as parameters, randomly initialized from standard normal distribution $\mathcal{N}(0,1)$. The hyper-parameters $\sigma_w$ and $\sigma_b$ control the scale of the weights and biases. For DNNs, we also require the stable conditions for $\sigma_w$ and $\sigma_b$ such that activations are bounded under $L\rightarrow \infty$. For ReLU network, we have stable configuration that $\sigma_w^2/2+\sigma_b^2=1$ , equivalently $\mathbb{E}(\|h^l\|)=\|x\|, \forall  1 \le l\le L$. We denote the output of an FCNN at time $t$ as $F_{\theta_t}(x)\equiv h^L$ with parameter set $\theta=\{\mathbf{W}^l,b^l\}_{0 \le l< L} $. The loss function measuring the discrepancy between prediction and the true label, is denoted as $\iota (h^L,y): \mathbb{R}^{n_L}\times \mathbb{R}^{n_L}\rightarrow \mathbb{R}$, where $y$ is the label for $x$. On training set $\mathcal{D}=\{(F_{\theta_t}(x),y)|x\in \mathcal{X}\} $, training process minimizes the loss values $\mathcal{L}=\mathbb{E}_{\mathcal{D}}[\iota (x,y)]$. In the following analysis, we normalize the input data such that $\|x\|=1$.






\subsection{Neural Network Gaussian Process Kernel}\label{sec.NNGPK}
As the width $n_l, \ l=1,2,...,L$ asymptotically goes to infinite, we conclude that the activations in each layer are i.i.d. zero-mean Gaussian random variables, whose behaviors are governed by the covariance matrix. As such, the evolution of activations through layers are approximated as Gaussian process and the correlation coefficient with given pair of data is called neural network Gaussian Process (NNGP)  kernel\footnote{In other papers, this kernel may be referred to as dual kernel \cite{daniely2016dualview} or activation kernel \cite{jacot2019freeze}.} \cite{xiao2019disentangling}. We define random vectors $u,v$ from probability space $\Omega_n(\rho) $ (in the scalar case where $n=1$, we will ignore the subscript), where $u,v$ are $n$-dimension Gaussian random variables with covariance value $\rho$, such that $(u,v) \sim \mathcal{N}(\mathbf{0},\left(
\begin{smallmatrix}
1 & \rho \\ \rho & 1 \end{smallmatrix}
\right) \otimes \mathbf{I_n})$.

Given two input data $x$ and $x'$ with covariance $\mathcal{K}^0(x,x')=\frac{1}{n_0}x^Tx'$, the kernel value in the $l$-th layer is calculated in a recursive manner.  

\begin{align}\label{eq.transop}
    \mathcal{K}^{l+1}&(x,x')\equiv\mathcal{T}(\mathcal{K}^{l}(x,x'))=\frac{\sigma_w^2}{n_ln_{l+1}} \\ &\mathbb{E}_{(u,v)\sim \Omega_{n_{l}}(\mathcal{K}^{l}(x,x'))}\left(    \phi(u)^T{\mathbf{W}^l}^T\mathbf{W}^l\phi(v) \right)+\sigma_b^2, \nonumber
\end{align}
where parameter $\mathbf{W}^l$ is initialized as Gaussian random matrix.
Previous papers \cite{jacot2018ntk,pen2017isometry} consider the infinite width scenario approximate that $\frac{1}{n_l}A^l\rightarrow\mathbf{I}_{n_l}$  and derive the relation:
\begin{equation}\label{eq.dualact}
    \mathcal{K}^{l+1}(x,x')=\sigma_w^2 \hat{\phi}(\mathcal{K}^{l}(x,x'))+\sigma_b^2,
\end{equation}
where $\hat{\phi}(\rho)\equiv \mathbb{E}_{(u,v)\sim \Omega(\rho)}( \phi(u)\phi(v) )$ is the dual activation function defined on $[-1,1]$. 
It tells how the activation function changes the covariance value. 
\citet{cho2009kernel} derives the closed-form dual activation functions for ReLU: 
\begin{equation}\label{eq.dualrelu}
    \hat{\phi}(\rho)=\frac{\sqrt{1-\rho^2}+(\pi-\arccos(\rho))\rho}{2\pi}.
\end{equation}
The transition operator $\mathcal{T}$ is fundamental in analysing dynamics of  networks. Stable initialization condition is expressed as $\mathcal{T}(1)=1$, implying activations of all layers have the same variance. \citet{daniely2016dualview} prove that  
there exists one and only one stable fixed point of $\mathcal{T}$. 
For ReLU, this fixed point is $\rho^*=1$.

\subsection{Neural Tangent Kernel}
Considering the gradient descent method to train neural networks, the derivative with parameters is $\nabla_\theta\mathcal{L}=(\nabla_\theta F_\theta(\mathcal{X}))^T\nabla_F\mathcal{L} $, and such 
gradient flow of network follows the differential equation at time $t$:
\begin{equation}\label{eq.gradflow}
    \frac{dF_{\theta_t}}{dt}=-\nabla_\theta F_{\theta_t}(\mathcal{X}) \nabla_\theta\mathcal{L}=-\Theta_t(\mathcal{X},\mathcal{X})\nabla_F\mathcal{L}.
\end{equation}
The kernel $\Theta_t=\sum_{p=1}^{|\theta|}\nabla_{\theta_p} F_{\theta_t}(\mathcal{X})\nabla_{\theta_p} F_{\theta_t}(\mathcal{X})^T$ is called \emph{Neural Tangent Kernel} (NTK), which maps a pair of data into an $n_L\times n_L$ matrix: $\mathbb{R}^{n_0}\times \mathbb{R}^{n_0} \rightarrow \mathbb{R}^{n_L\times  n_L}$.

The key conclusion \cite{jacot2018ntk,arora2019cntk,bietti2019inductive} of NTK is its convergence behavior as the layer width goes to infinity, which converges in probability to a constant kernel:
\begin{equation}\label{eq.constntk}
    \Theta_t\rightarrow\Theta\otimes I_{n_L},
\end{equation}
where the scale kernel is calculated recursively by previous NNGP kernel:
\begin{equation}\label{eq.ntk}
    \Theta(x,x')=\sum_{l=1}^L\mathcal{K}^l(x,x')\prod_{r=l+1}^L\dot{\mathcal{T}}(\mathcal{K}^r(x,x')),
\end{equation}
where $\dot{\mathcal{T}}$ denotes the derivative of $\mathcal{T}$.
Since the NTK describes the dynamics for neural network during training, the condition number of NTK implies the convergence of neural network. \citet{lee2019linearmodels} prove that the NTK dynamics is linear approximation of neural network with initialization. Despite Equation~(\ref{eq.constntk}) is derived under an infinite width, the condition number of NTK is still an effective criteria indicating the training process.

\section{Equivalence of Weight Mean and BN}\label{sec.wm}
\begin{figure}[t]
  \centering
  \includegraphics[width=0.23\textwidth]{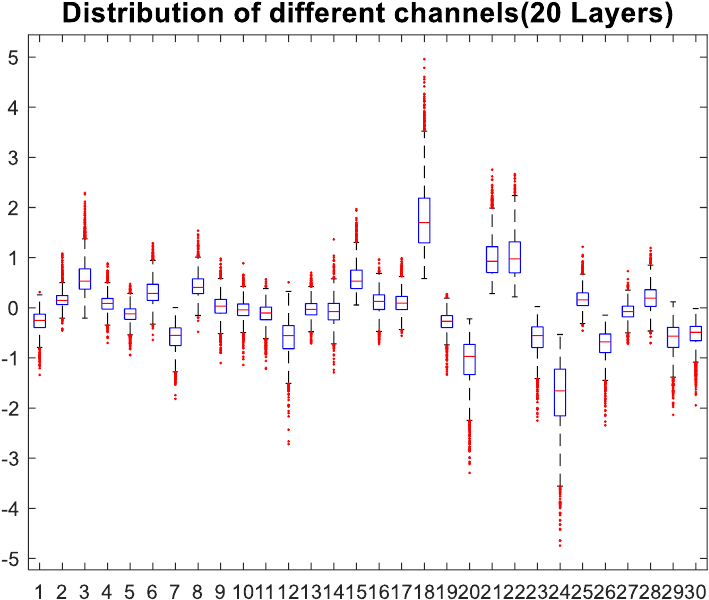}   
  \includegraphics[width=0.23\textwidth]{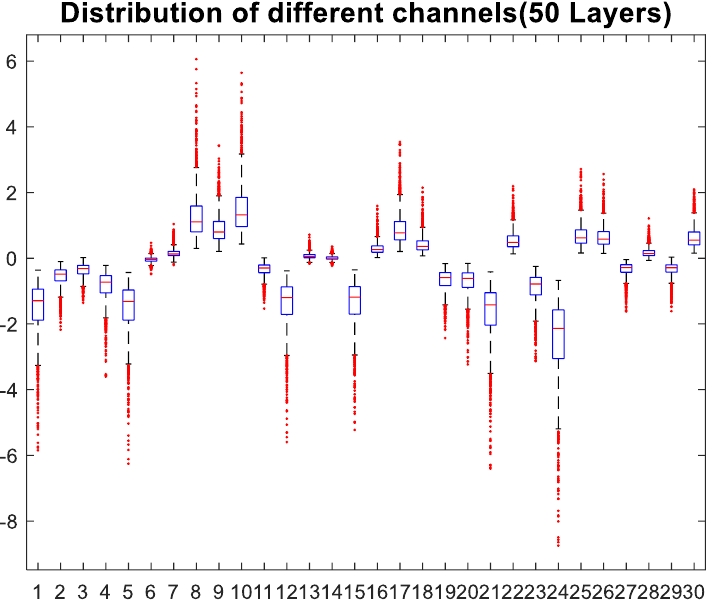}
  \caption{Straight network correlates activations. We randomly select 30 channels in the 20-th and 50-th layers, and show their distributions at initial iteration. Notice that most values are located within boxes, we conclude that the intra-channel similarities and inter-channel variations increase with depth. In other words, different inputs generate similar activations. This correlated activations are results of the frozen NNGP kernel.}
  \label{fig.shiftmean}
\end{figure}

It is well known that whitening input is beneficial to accelerating and improving learning algorithms. However, input whitening is insufficient on deep learning methods, as the distributions of activations vary with layers. BN is designed to reduce those variations, \emph{i.e.}, internal covariate shift (ICS). From the perspective of kernel method, the ICS refers to the frozen NNGP kernel. Figure~\ref{fig.shiftmean} illustrates the assorted distributions on different layers and channels. We test a 51-layer FCNN with ReLU activation that is initialized as $(\sigma_w,\sigma_b)=(\sqrt{2},0)$ with batchsize 2000, and plot their distributions on very beginning iteration. The intra-channel values concentrate while inter-channel values disperse with the growth of depth. When the depth increases to infinity, network generates constant values, and NNGP kernel freezes.
 

\subsection{Fundamental Impact of BN} 
In this part, we will give the explanation for the frozen NNGP kernel in Figure~\ref{fig.shiftmean} and show how BN layers revise this kernel. Recall the definition of transition operator $\mathcal{T}$ in Eq.~\ref{eq.transop} describing the relationship of correlation coefficients between adjacent layers.
For straight network, the only fixed point of $\mathcal{T}$ is $\rho^*=1$, which indicates that the different input will correlate as layer increases,
\begin{equation}
    \lim_{l} \mathcal{K}^{l}(x,x')= \rho^*=1\quad  \forall x,x' \in \mathcal{X}.
\end{equation}
This frozen NNKG kernel also dominates the value of NTK (see Equation~(\ref{eq.ntk})) and hinders convergence of network. \citet{xiao2019disentangling} characterize the network trainability by analyzing the spectrum of NTK. They divide the network training dynamic into three different phases according to the value $\chi_1=\dot{\mathcal{T}}(1) $, and prove the following theorem.
\begin{theorem}\label{theorem1}[\cite{xiao2019disentangling}]
 The limiting condition number $\kappa$ of NTK is characterized into three parts: 1) the ordered phase with $\chi_1<1$ has divergent $\kappa$, 2) the critical line with $\chi_1=1$ has condition number linear to training data-size $\kappa\sim|\mathcal{X}|$, 3) the chaotic phase with $\chi_1>1$ has constant limitation that $\kappa\rightarrow 1$. 
\end{theorem}
 
In over-parameterized networks, the condition number of NTK controls the trainability. Only in chaotic phase, network can achieve linear convergent rate. For straight ReLU networks, the stable initialization leads to large condition numbers, such that its training process is usually troubling.

\begin{proposition}\label{prop.bn}
  With definitions in Theorem~\ref{theorem1}, the NTK of a deep FCNN is in ordered phase or critical line with $\chi_1\le 1$, where equality holds when parameters are initialized with $(\sigma_w,\sigma_b)=(\sqrt{2},0)$. 
\end{proposition}

Note that the derivative of dual activation in Equation~(\ref{eq.dualrelu}) is 
$\mathrm{d}\hat{\phi}(1)/\mathrm{d}\rho=1/2$, combining this fact with Equation~(\ref{eq.dualact}) derives this proposition. The critical initialization $(\sigma_w,\sigma_b)=(\sqrt{2},0)$ is the default setting for ReLU network as it stabilizes gradient propagation without vanishing or exploding \cite{pen2017isometry}. However, this initialization also freezes the NNGP kernel and leads to nearly singular NTK. 

Besides the slow convergence rate, the straight network also leads to inferior accuracy. An intuitive explanation comes from lower network expressivity. As network outputs similar value on each channel, as shown in Figure~\ref{fig.shiftmean}, the channels with positive value behave like linear unit and negative channels are totally shut down. Therefore, straight ReLU network is more like a narrower linear system, which is incapable to fit complex functions.

Furthermore, we show that BN networks are in chaotic regime. We study the NNGP kernel and NTK in BN network. Batch normalization whitens intermediate activations to avoid the frozen NNGP kernel. We consider the common setting where BN layer locates behind FC layer. For simplicity, we consider training with full-batch and the BN operator is defined as:
\begin{equation}
    BN(h^l)=\frac{h^l-m^l}{\sqrt{\nu^l}},\ m^l=\mathbb{E}_{\mathcal{X}}[h^l(x)],\nu^l=\mathbb{E}_{\mathcal{X}}(h^l(x)-m^l)^2.
\end{equation}

After BN layer, each channel is normalized to have the same distribution with zero mean and unit variance. As a result, the outputs of BN networks are expected to decorrelate and the NNGP kernel tends to an identical matrix.

\begin{theorem}\label{the.bn}
  For a deep FCNN with BN layers following each FC layer, the criterion converges in probability that $\chi_1 \overset{p.}{\rightarrow} \frac{1}{1-1/\pi}$ as layer width goes to infinity, such that BN networks are in chaotic regime. 
\end{theorem}
Please refer to Appendix~A for proof. BN is believed to reduce the ICS that mathematically refers to unfreezing the NNGP kernel in our opinion. Theorem~\ref{the.bn} implies that one of the fundamental benefit from BN layer is to reduce the condition number of NTK and improve convergence rate. 

\begin{figure}[t]
  \centering
  \includegraphics[width=0.23\textwidth,height=0.18\textwidth]{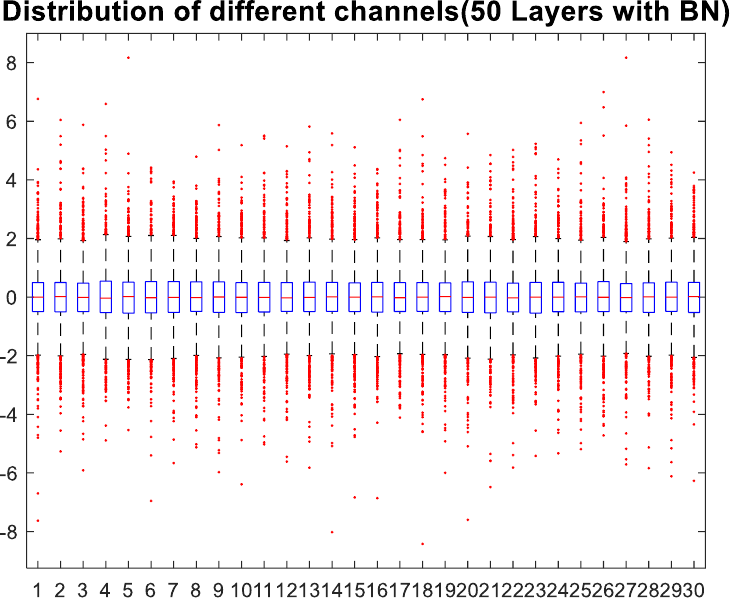}  
  \includegraphics[width=0.23\textwidth,height=0.18\textwidth]{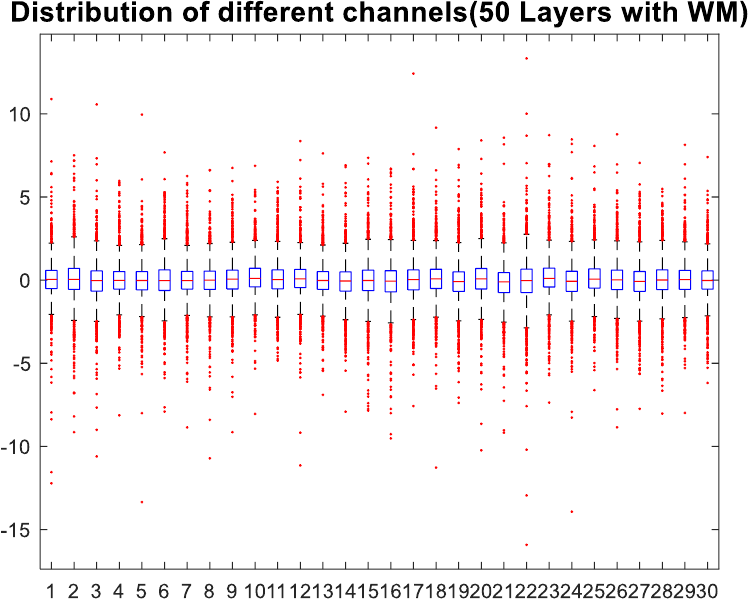}
  \caption{BN layer and weight mean operation unfreeze NNGP kernel. This figure plots activation distributions in the 50-th layers with BN layers (left) and weight mean operation (right). Activation in each channel follows the same distribution.}
  \label{fig.bnwm}
\end{figure} 

\subsection{Transition Operation by Weight Mean}
Our weight mean operation subtracts the average value along each output channel. Moreover, we also rescale the weight parameters by multiplying $1/\sqrt{1-1/\pi}$ to satisfy the stable condition. This scaling compensates the variance decrease as we subtract mean values along each channel. We firstly show the experimental results to prove that the weight mean works similarly as BN layers. Figure~\ref{fig.bnwm} exhibits distributions on networks with BN layers and our weight mean operation. Both two methods lessen the inter-channel variation and decorrelate data. 

To analyze the mechanism of mean operation, We consider weight matrix $\mathbf{W}$ with shape $(n_o,n_i)$ and denote its post-mean matrix as $\bar{\mathbf{W}}=\mathbf{W}-w\mathbf{1}^T$, where $w$ is an $n_o$-dimension vector. Following the definition in Equation~(\ref{eq.transop}), the transition operator $\bar{\mathcal{T}}$ is calculated by:
\begin{equation}
    \bar{\mathcal{T}}(\rho)=\frac{\sigma_w^2}{n_in_o}  \mathbb{E}_{(u,v)\sim \Omega_{n_i}(\rho)}\left(    \phi(u)^T\bar{\mathbf{W}}^T\bar{\mathbf{W}}\phi(v) \right)+\sigma_b^2. 
\end{equation}

There is an intuitive explanation about how weight mean reduces correlation. Since each component of weight matrix is initialized as Gaussian random variable, the mean vector $w$ is also Gaussian with variance scaled by layer width $n$. In ReLU network, activations are clipped to be non-negative values, such that the input of FC layer $\phi(u)$ contains large positive component. As a result, weight mean diminishes a positive value $w\mathbf{1}^T\phi(u)$ for every channel and decorrelate data. For transition operator, we have the following results.

\begin{lemma}\label{lemma}
  For ReLU activation $\phi(x)=max(x,0)$, and define $\Phi(\rho)=\mathbb{E}_{(u,v)\sim \Omega_{n_i}(\rho)}\left(\phi(u)^T\mathbf{W}^T\mathbf{W}\phi(v) \right),\ \bar{\Phi}(\rho)=\mathbb{E}_{(u,v)\sim \Omega_{n_i}(\rho)}\left(\phi(u)^T\bar{\mathbf{W}}^T\bar{\mathbf{W}}\phi(v) \right)$, the following equation holds
  \begin{equation}\label{eq:l1}
      \mathbb{E}_W[\bar{\Phi}(\rho)]=\frac{n_i-1}{n_i}\mathbb{E}_W[\Phi(\rho)-\Phi(0)].
  \end{equation}
\end{lemma}

The proof of Lemma~\ref{lemma} is provided in Appendix~A. The norm of output is reduced by subtracting a constant value. From Equation~\eqref{eq:l1}, we quantify the decline of norm of activations. In general configuration of deep networks, bias terms are initialized to zero (\emph{i.e.}, $\sigma_b=0$). We rescale the initial variance of weight $\sigma_w$ to achieve stable condition $\bar{\mathcal{T}}(1)=1$.

\begin{figure}[t]
  \centering
  \includegraphics[width=0.35\textwidth]{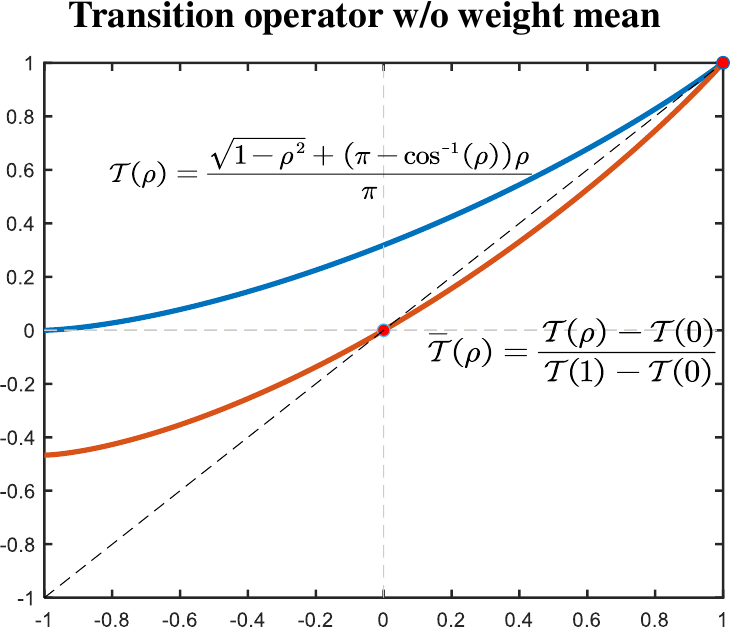}
  \caption{Transition operator for networks with and without weight mean operation. Operator $\mathcal{T}$ for straight network (blue curve) has stable fixed point at $\rho^*=1$ indicating correlated inputs and the frozen NNGP kernel. While the stable fixed point of operator $\bar{\mathcal{T}}$ (orange curve) locates at $\rho^*=0$ and mean weight turns network into chaotic regime as the criterion $\chi_1=\dot{\bar{\mathcal{T}}}(1)>1$. }
  \label{fig.opplot}
\end{figure} 

\begin{theorem}\label{the.wm}
  Considering zero-bias configuration, 1) straight deep ReLU networks are stably initialized by $\sigma_w^2=2$, and have transition operator $\mathcal{T}(\rho)=2\hat{\phi}(\rho)$ with $\chi_1=1$; 2) deep ReLU networks with weight mean are stably initialized by $\sigma_w^2=\frac{2n}{(n-1)(1-1/\pi)}$ where $n$ is the input dimension, and have $\bar{\mathcal{T}}(\rho)=\frac{ \hat{\phi}(\rho)-\hat{\phi}(0)}{\hat{\phi}(1)-\hat{\phi}(0)}$ with $\chi_1=\frac{1}{2(\hat{\phi}(1)-\hat{\phi}(0))}=\frac{1}{1-1/\pi}$.
\end{theorem}

Figure~\ref{fig.opplot} plots the function relationships and fixed points of transition operators with and without weight mean. Weight mean changes the fixed stable point from $\rho^*=1$ to $\rho^*=0$, which indicates that input data decrease their correlation as depth increases and the input data become decorrelated. As for NTK, the criterion $\chi_1$ with weight mean is identical to BN networks and implies chaotic kernel. Therefore, networks with weight mean have the same convergence rate as BN layers.

\section{Implementation of MimicNorm}\label{sec.mimicnorm}
\begin{wrapfigure}{r}{3.0cm}
\includegraphics[width=2.6cm]{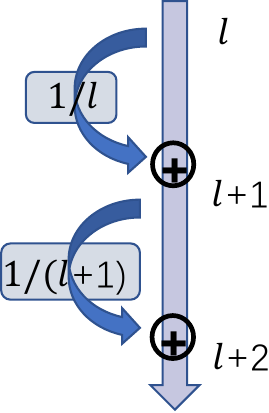}
  \caption{Batch Normalization suppresses variance in $l$-th residual branch by a factor $1/l$.}
  \label{fig.resdual}
\end{wrapfigure}

In this paper, we propose \emph{MimicNorm} to replace BN layers and reduce memory consumption. There are also some potential benefits from eliminating intermediate BN layers, such as stable training for micro-batch scenario. The main components of \emph{MimicNorm} are weight mean for convolutional layers and one last BN layer, as illustrated in Figure~\ref{fig.mimic}. 
 
\subsection{Weight Mean Operation}
Our weight mean operation subtracts mean values along each channel, similar to Weight Standardization \cite{qiao2019weight}. Besides, we modify the initialization according to Theorem~\ref{the.wm}, with a scaling factor of $\frac{1}{\sqrt{1-1/\pi}}\approx 1.2$ in straight forward structure. 
As for skip-connection, we add one scalar multiplier in the end of each residual branch, similar to SkipInit \citet{de2020batch}. We initialize this scalar to generate normalized activations. Considering the residual network illustrated in Figure~\ref{fig.resdual}, the input of the $l$-th residual block is expected to have accumulated variance $l$. And to normalize the output, this residual branch implicitly downscales variance by $l$. In this way, the learnable scalar has an initial value $1/\sqrt{l}$. Our method also works well for lightweight networks like ShuffleNet, which contains depthwise convolutional layers. Our weight mean ignores those layers, since the total number (depending on kernel size, usually $3\times3=9$) of weights in one channel is too small.


\subsection{Last BN Layer}
Empirical results have shown that the straight networks are only trainable with smaller learning rates \cite{bjorck2018understanding}.
We observe that a larger learning rate usually causes output explosion within several iterations. We conjecture that it is the gradient that encourages variance of outputs. An intuitive explanation is that two contradictory forces control the evolution of variance, namely scattering inter-class data and congregating intra-class ones. A larger learning rate weights the scattering effect and may lead to divergence. BN layer auto-tunes the learning rates, as backpropagated gradients shrink by variance. However, this auto-tuned learning rate can be implemented by one last BN layer. Therefore, the unit variance operations in former layers are unnecessary. 

As for implementation, our last BN layer locates after the finial fully-connected layer and normalizes $(B,C)$ output matrix, where $B$ is batchsize and $C$ is the number of categories. Compared with former BN layers, the last one is much lighter and requires less resources. Besides, we also disable affine parameters to keep the same distribution of classification categories. 

 \section{Experiments}\label{sec.exp}
We employ MimicNorm on various networks and compare it with original BN networks and other implementation without BN layers, \emph{i.e.} Fixup initialization \cite{zhang2019fixup}. Fixup method initializes classification and residual branch to be zero, downscales weight parameters in convolutional layers, and induces scalar biases (and multiplier) before and after convolutional layers. They avoid divergence by repressing output to be zero and succeed in training residual networks without BN layers. Our experiments are evaluated in CIFAR-100 dataset and ImageNet dataset. We also do an ablation study to investigate the isolated impact. Besides, some empirical discussions about weight mean and last BN layer are attached in Appendix~B. Experiments in this paper are implemented  with Pytorch framework, version 1.5, and code is available at https://github.com/Kid-key/MimicNorm.
\subsection{Evaluations on CIFAR-100}
In this section, we show results of our method and BN networks on CIFAR-100 dataset, which are evaluated by one GPU with 12GB memory, GeForce RTX 2080 Ti. Our optimizer is firstly warmed up within 2 epochs from 0 to 0.1, and then uses multi-step scheduler decayed by 0.1 at epochs [80,110,135]. The default batchsize is 256. However, ResNet50 requires more memory to deploy, while we halve the batchsize into 128. Table~\ref{tab.exp1} presents the accuracy and memory consumption on straight forward networks (VGG11/16 \cite{simonyan2014very}), skip connected networks (ResNet18/50 \cite{he2016deep}) and lightweight networks (SqueezeNet \cite{hu2018squeeze}, ShuffleNetV2 \cite{ma2018shufflenet}).

\begin{table}[!t]
\centering
\begin{threeparttable}
\begin{tabular}{llll} 
\toprule
Model                  & Method    & Accuracy (\%)            & Memory  \\ 
\hline
\multirow{3}{*}{VGG11} & BatchNorm & $67.90\pm 0.14$ & 2971MB  \\ 
 
                        & \textbf{MimicNorm} & $67.87\pm 0.20$ & 2851MB  \\ 
 
                       & NoNorm & $65.33\pm 0.08$ & 2831MB  \\ 
\hline
\multirow{3}{*}{VGG16} & BatchNorm & $72.28\pm 0.05$ & 3745MB  \\ 
 
                       & \textbf{MimicNorm} & $72.47\pm 0.02$ & 3489MB  \\ 
                       
                       & NoNorm & $68.97\pm 0.36$\tnote{*} & 3469MB  \\ 
\hline
\hline
\multirow{3}{*}{ResNet18} & BatchNorm & $74.72\pm 0.31$ & 4447MB  \\ 
 
                       & Fixup & $70.48\pm 0.18$ & 4743MB  \\  
                       
                       & \textbf{MimicNorm} & $74.96\pm 0.14$ & 4157MB  \\  
\hline
\multirow{3}{*}{ResNet50} & BatchNorm & $78.53\pm 0.26$ & 8533MB  \\ 
 
                        & Fixup & $72.81\pm 0.10$\tnote{*} & 10283MB  \\  
                        
                       & \textbf{MimicNorm} & $77.91\pm 0.20$ & 7533MB  \\ 
\hline
\hline
\multirow{2}{*}{SqueezeNet} & BatchNorm & $70.88\pm 0.08$ & 3687MB  \\ 
 
                       & \textbf{MimicNorm} & $71.15\pm 0.10$ & 2693MB  \\ 
\hline
\multirow{2}{*}{ShuffleNetV2} & BatchNorm & $71.10\pm 0.49$ & 3915MB  \\ 
 
                       & \textbf{MimicNorm} & $70.93\pm 0.13$ & 3141MB  \\                
\bottomrule
\end{tabular}
\begin{tablenotes}
       \footnotesize
       \item[*] training is unstable, and fails to converge sometimes.
\end{tablenotes}
\end{threeparttable}
\caption{Comparison with various network structures on CIFAR-100 dataset. We evaluate
 three times and calculate mean and std of accuracy. On all structures, our method achieves comparable accuracy and beneficial memory consumption.}
 \label{tab.exp1}
\end{table}

For straight structures, our method exhibits the same impacts of BN layers, \emph{i.e.} improving accuracy and stabilizing training process. VGG16 network may divergence with learning rate $0.1$, and both our method and BN network are stable. We observe that our method only requires 20MB more memory, which demonstrates the low complexity of the last BN layer. For the ResNet structures, we also compare with Fixup initialization, which is able to train deep ResNets without normalization. Our method saves about 10\% memory with similar accuracy. While Fixup initialization dramatically deteriorates the performance and increases memory consumption. Despite Fixup initialization removes all BN layers, this method induces learnable scalar parameters. To calculate the gradients of scalar parameters, Pytorch implementation stores lots of temporary tensors consuming huge memory. We also validate our method on lightweight networks, which are more sensitive to hardware resources. Our method obtains more than 20\% memory saving with the same results. To the best of our knowledge, our method is the first attempt to train lightweight networks without intermediate BN layers.  

\subsection{Ablation Studies}
We further separately evaluate the impact of weight mean and the last BN layer. As illustrated in previous sections, weight mean operation unfreezes the NNGP kernel and accelerates training process. The last BN layer auto-tunes learning rate to avoid exploded variance. 

\begin{figure}[!t]
  \centering
  \includegraphics[width=0.36\textwidth]{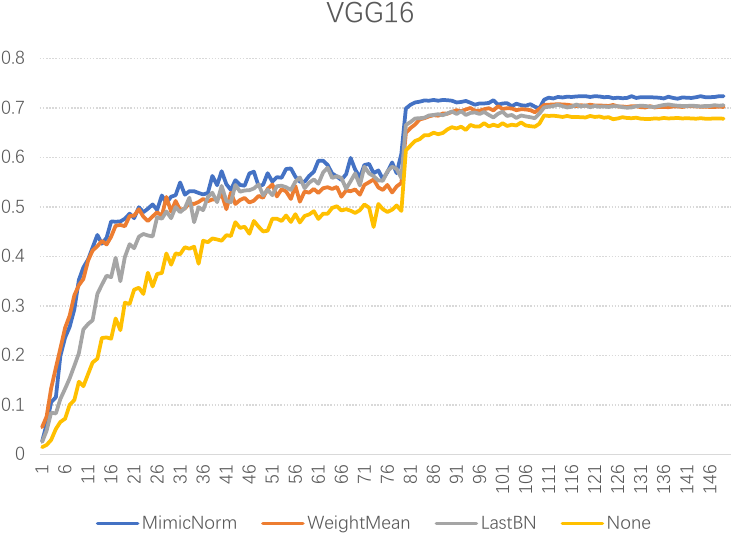} 
  \caption{Ablation study on VGG16 network. we observe that both weight mean and last BN layer contribute to the improvement of performance. while weight mean accelerates training at the initial phrase. }
  \label{fig.vggabl}
\end{figure} 

Figure~\ref{fig.vggabl} compares the training processes of different methods on VGG16. In the beginning, weight mean enables faster training, in accordance with the theoretical prediction. The last BN layer also accelerates training and improves accuracy, while our method combining weight mean and the last BN layer achieves the best accuracy. Table~\ref{tab.vgglr} evaluates the stability of the last BN layer, where all the methods are test using different learning rates. For small learning rates, all the methods succeed in training and weight mean achieves faster convergence. For large learning rates, last BN layer is essential for training. Moreover, Figure~\ref{fig.vggabl} shows our method yields a larger accuracy within the first 10 epochs in comparison to BN implementation.

\begin{table}
\centering
\begin{tabular}{l|rrrr}
\toprule
\begin{tabular}[c]{@{}l@{}}Learning \\Rate\end{tabular} & \multicolumn{1}{l}{None} & \multicolumn{1}{l}{\begin{tabular}[c]{@{}l@{}}Weight\\Mean\end{tabular}} & \multicolumn{1}{l}{\begin{tabular}[c]{@{}l@{}}Mimic\\Norm\end{tabular}} & \multicolumn{1}{l}{\begin{tabular}[c]{@{}l@{}}Batch\\Norm\end{tabular}}   \\
\hline
0.01           & 22.12      & 34.81           & 38.39        & 37.82                           \\
0.02           & 26.76      & 43.07           & 44.32        & 40.49                           \\
0.05           & 25.63      & 42.63           & 49.53        & 37.97                           \\
0.1            & 1.24       & 36.49           & 46.92        & 32.4                            \\
0.2            & 1.77       & 1.00            & 39.41        & 25.5                            \\
0.5            & 1.00       & 1.00            & 22.03        & 8.26                            \\
1.0            & 1.00       & 1.00            & 9.83         & 3.93       \\          \bottomrule    
\end{tabular}
\caption{Best accuracy (\%) on the first 10 epochs with different learning rate. The last BN layer makes network trainable among larger learning rates.}
\label{tab.vgglr}
\end{table}
\begin{figure*}[t]
  \centering
  \includegraphics[width=0.36\textwidth]{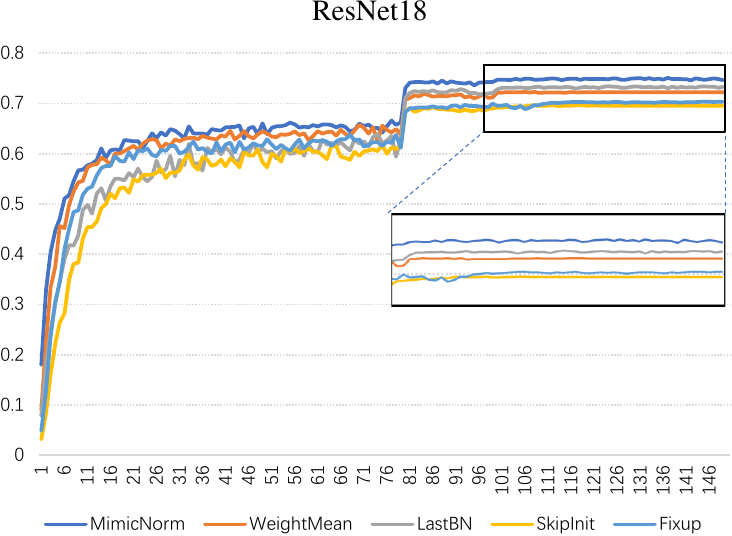} \quad \quad
  \includegraphics[width=0.36\textwidth]{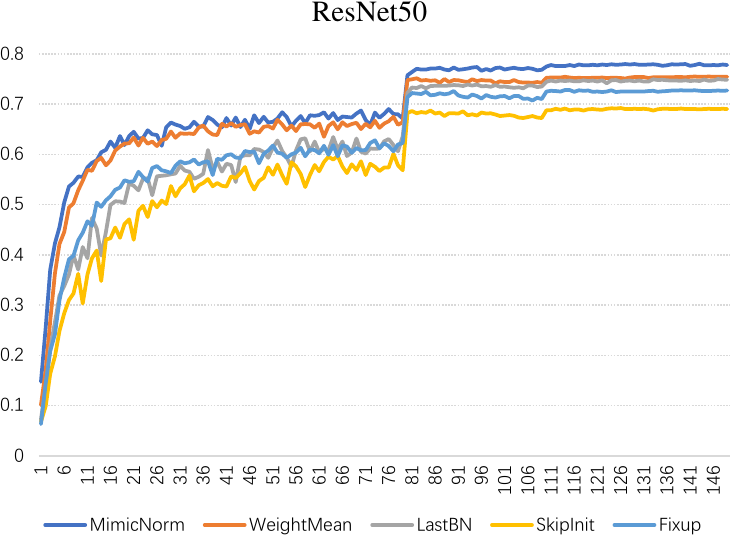}
  \caption{Comparison of training curves on ResNet18 and ResNet50. Weight mean accelerates training at the beginning and last BN layer improves accuracy. Besides, our method also gains large superiority over Fixup initialization. }
  \label{fig.resabl}
\end{figure*} 
Despite our analysis is based on straight forward structures, the empirical results show that weight mean also accelerates training for ResNets. Ablation results on Resnet18 and Resnet50 are presented in Figure~\ref{fig.resabl}. 
Note that our method adds a scalar at the end of each residual branch, and initializes it according to SkipInit method. 
Our method surpasses state-of-the-art methods, with faster training and better accuracy. We also observe one last BN layer in ResNet18 generates better results than weight mean. However, for deeper network, one last BN layer shows inferiority than weight mean because of the much worse performance at the initial phase. Thus, weight mean is more important for deep networks as it avoids the frozen NNGP kernel, and one last layer is able to compensate most accuracy drop for shallow networks.

\subsection{Evaluations on ImageNet}
On ImageNet dataset, we experiment on Tesla V100 GPUs. We follow exactly the same data augmentation strategy as official Pytorch implementation: random flipping and sampling, and resizing to $224 \times 224$ images. The training optimizer is also general SGD optimizer with momentum factor 0.9 and weight decay 1e-4. We train network for 90 epochs with the first 2 epochs warming up and decrease learning rate from 0.1 to 0.0001 at milestones [30, 60, 85]. 
\begin{table}[!t]
\centering
\begin{threeparttable}
\begin{tabular}{llll} 
\toprule
Model                  & Method    &  \multicolumn{1}{l}{\begin{tabular}[c]{@{}l@{}}Accuracy\\(\%)\end{tabular}}            & \multicolumn{1}{l}{\begin{tabular}[c]{@{}l@{}}Total\\memory\end{tabular}}  \\ 
\hline
\multirow{3}{*}{VGG16\tnote{*}} & BatchNorm & $73.89$ & 62162MB  \\ 
 
                       & \textbf{MimicNorm} & $73.31$ & 56410MB  \\ 
                       
                       & NoNorm & $71.59$ & 56388MB  \\ 
\hline
\multirow{2}{*}{ResNet18} & BatchNorm & $69.76$ & 8057MB  \\ 
                      
                       & \textbf{MimicNorm} & $70.15$ & 5603MB  \\  
\hline
\multirow{3}{*}{ResNet50} & BatchNorm & $76.48$ & 27083MB  \\ 
 
                        & Fixup & $72.40$ & 29267MB  \\  
                        
                       & \textbf{MimicNorm} & $76.84$ & 22489MB  \\ 
\hline
\multirow{2}{*}{ResNet101\tnote{*}} & BatchNorm & $77.37$ & 37168MB  \\ 
                      
                       & \textbf{MimicNorm} & $77.52$ & 30363MB  \\  
\hline
\multirow{2}{*}{\begin{tabular}[c]{@{}l@{}}ShuffleNet\\V2\_0.5\end{tabular}} & BatchNorm & $59.77$ & 4385MB  \\ 
 
                       & \textbf{MimicNorm} & $58.31$ & 3603MB  \\                
\bottomrule
\end{tabular}
\begin{tablenotes}
       \footnotesize
       \item[*] Model deployment with batchsize 256 requires 2 GPUs.
\end{tablenotes}
\end{threeparttable}
\caption{Test results on ImageNet dataset with batchsize 256. Results show our method gets comparable accuracy with BN layers, and exceeds Fixup method largely. Note for ResNet50/110 and ShuffleNetV2, Our method saving nearly 20\% memory.}
\label{tab.exp2}
\end{table}

Table~\ref{tab.exp2} shows the results on several network structures. Similar to our finding on CIFAR-100 dataset, our method has similar accuracy on all structures and saves much more memory. We save about 10\% memory for VGG16, less than ResNets and lightweight structures, because the VGG nets have less BN layers and larger size of feature maps. On ResNet structures, we achieve impressive performance, higher accuracy than BN networks and about 20\% memory saving. In comparison, Fixup initialization is inferior both in accuracy and memory. As for ShuffleNetV2, we modify the training strategy according to the original paper \cite{zhang2018shufflenet} as the results under our default setting is much less than claimed accuracy, both for our method and BN implementation. We use less data augmentation and train for 200 epochs. Our method slightly underperforms BN implementation, which may be a result of depthwise convolutions. Note that our method skips weight mean for depthwise convolutional layers and there is no regularization for them. Despite the slight deficit, we save nearly 20\% total memory even for lightweight network structure.

 \section{Conclusion and Discussion}

In this paper, we propose MimicNorm to replace BN layers and reduce memory consumption, which consists of two heart components, weight mean and last BN layer. We theoretically demonstrate that, in an infinite width DNN, weight mean and BN layers unfreeze the NNGP kernel leading to chaotic NTK, and enable linear convergence rate. For the finite width network, NTK describes the one-order approximation at initialization. Thus, networks with chaotic NTK converge faster at first several epochs. And empirical evidences confirm the theoretical prediction. On the other hand, last BN layer adjusts learning rate by the variance of output. We show last BN layer stabilizes training process on larger learning rates. Additionally, we also find that the auto-tuned learning rate leads to better performance and its dynamic is still unclear. In our implementation, last BN layer could be folded into loss function, which implies that a variance-aware loss function may facilitate network training.

\bibliography{MimicNorm}
\section*{Appendix. A: Proofs}
\subsection*{\textbf{Theorem 2.}}

\emph{ For a deep FCNN with BN layers following every FC layer, the criterion converges in probability that $\chi_1 \overset{p.}{\rightarrow} \frac{1}{1-1/\pi}$ as layer width goes to infinity, such that BN networks are in chaotic regime. }

\emph{Proof.} 
With Batch Normalization after FC layer, the bias term always keep zero such that $\sigma_b=0$. Besides, the intrinsic property of BN that insensitive to norm of parameter, our analysis choose initialization $\sigma_w=1$.  We have transition operator $\mathcal{T}$ in $l$-th layer as formula:
\begin{align}
    \mathcal{T}(\rho)&=\frac{1}{n_{l+1}}  \mathbb{E}_{(u,v)\sim \Omega_{n_{l}}(\rho)}\left( BN(h^{l+1}(u))^T BN(h^{l+1}(v))  \right) \nonumber \\
    &=\frac{1}{n_{l+1}} \sum_i^{n_{l+1}} \mathbb{E}_{(u,v)\sim \Omega_{n_{l}}(\rho)}\left( BN(h^{l+1}_i(u)) BN(h^{l+1}_i(v))  \right)
\end{align}
where, random vector $h^{l+1}(u)= \frac{\mathbf{W}^l\phi(u)}{\sqrt{n_l}}$ and $h^{l+1}_i$ is the $i$-th component. We will ignore the up script $l$ and $l+1$ if no ambiguity and define $n_o=n_{l+1},n_i=n_l$ (subscript $o/i$ means output/input dimension for this layer). Notice the definition of BN operator, the mean and variance value is expected to dataset, i.e. $m=\mathbb{E}_u[h(u)]$ and $\nu=\text{Var}_u[h(u)]$. We have:
\begin{align}\label{eq.ap2}
    \mathcal{T}(\rho)&=\frac{1}{n_o} \sum_i^{n_o} \mathbb{E}_{(u,v)\sim \Omega_{n_i}(\rho)}\left( BN(h_i(u)) BN(h_i(v))  \right) \nonumber \\
    &=\frac{1}{n_o} \sum_i^{n_o} \frac{1}{\nu_i} \mathbb{E}_{(u,v)\sim \Omega_{n_i}(\rho)}\left( (h_i(u)-m_i) (h_i(v)-m_i) \right) \nonumber \\
    &=\frac{1}{n_o} \sum_i^{n_o} \left[\frac{1}{\nu_i} \mathbb{E}_{(u,v)\sim \Omega_{n_i}(\rho)}( h_i(u) h_i(v) ) -m_i^2\right] \nonumber \\
    &=\frac{1}{n_o} \sum_i^{n_o} \left[\frac{1}{\nu_i} \hat{\phi}(\rho) - m_i^2\right]
\end{align}
The $\hat{\phi}(\rho)$ is the dual activatio of ReLU defined in Eq.~2, which dominates NNGP kernel of straight network. Notice $\frac{d}{d\rho}\hat{\phi}(1) = 1/2$, we
calculate its derivative at $\rho=1$, 
\begin{equation}\label{eq.ap1}
    \chi_1 = \frac{d}{d\rho}\mathcal{T}(1)=\frac{d}{d\rho}\frac{1}{n_o} \sum_i^{n_o} \left[\frac{1}{\nu_i} \hat{\phi}(1) - m_i^2\right] 
    =\frac{1}{2n_o} \sum_i^{n_o} \frac{1}{\nu_i} 
\end{equation}

We find the different on $\chi_1$ led by BN is the average variance term. However, $\chi_1$ is still random variable depending on Gaussian Random matrix $W$. We will prove as layer width approaching infinity, the variance converge in probability to a constant and valid our theorem.
\begin{align}
    \nu_i&=\text{Var}_u[h_i(u)]=\text{Var}_u[\frac{\mathbf{W}_i\phi(u)}{\sqrt{n_i}}] \nonumber \\
    &=\frac{1}{n_i}\text{Var}_u(\sum_j^{n_i}\mathbf{W}_{ij}\phi(u_j)) \nonumber \\
    &=\frac{1}{n_i}\sum_j^{n_i}\mathbf{W}_{ij}\text{Var}_u(\phi(u_j))
\end{align}
For ReLU function $\phi(x)=\max(x,0)$, we have 
$\mathbb{E}_u(\phi(u))=\sqrt{\frac{1}{2\pi}}$ and 
$\text{Var}_u(\phi(u))=\mathbb{E}_u(\phi(u)^2)-\mathbb{E}_u(\phi(u))^2=\frac{1}{2}-\frac{1}{2\pi}$
Denote $S=1-1/\pi$, such that
\begin{equation}
    \nu_i=\frac{S}{2n_i}\sum_j^{n_i}\mathbf{W}_{ij}
\end{equation}
Leveraging the Large Number Theorem, the average value of any raw of random matrix $W$ converges, i.e. $\nu_i \overset{p.}{\rightarrow} \frac{S}{2}, \forall\  i$. Combining Eq. \ref{eq.ap1}, we get our conclusion.
\begin{equation}
    \chi_1 =\frac{1}{2n_o} \sum_i^{n_o} \frac{1}{\nu_i} \overset{p.}{\rightarrow} \frac{1}{2n_o} \sum_i^{n_o} \frac{2}{S}=\frac{1}{S}
\end{equation}

\subsection*{\textbf{Lemma 1.}}

  \emph{For ReLU activation $\phi(x)=max(x,0)$, and define $\Phi(\rho)=\mathbb{E}_{(u,v)\sim \Omega_{n}(\rho)}\left(\phi(u)^T\mathbf{W}^T\mathbf{W}\phi(v) \right),\ \bar{\Phi}(\rho)=\mathbb{E}_{(u,v)\sim \Omega_{n}(\rho)}\left(\phi(u)^T\bar{\mathbf{W}}^T\bar{\mathbf{W}}\phi(v) \right)$, the following equation holds.}
  \begin{equation}
      \mathbb{E}_W[\bar{\Phi}(\rho)]=\frac{n_o-1}{n_o}\mathbb{E}_W[\Phi(\rho)-\Phi(0)] \nonumber
  \end{equation}
\emph{Proof.} The different from $\Phi(\rho)$ and $\bar{\Phi}(\rho)$ is the symmetric random matrix. We denote $\mathbf{A}=\mathbf{W}^T\mathbf{W}$ and $\bar{\mathbf{A}}=\bar{\mathbf{W}}^T\bar{\mathbf{W}}$ and investigate their element wise distribution.
For $\mathbf{A}$, the diagonal component $\mathbf{A}_{ii}$ subjects to Chi-Square distribution that
\begin{equation}
    \mathbf{A}_{ii}=\mathbf{W}_{i}^T\mathbf{W}_{i}\sim \chi^2(n_o)
\end{equation}
while, due to the independent initialization of $W$, the non-diagonal elements  has expectation that $\mathbb{E}[\mathbf{A}_{ij}]=0, \forall\ i\ne j$. Now, we calculate as
\begin{align}
    \mathbb{E}_{W}[\Phi(\rho)]&=\mathbb{E}_{W}\mathbb{E}_{(u,v)}(\phi(u)^T\mathbf{A}\phi(v)) \nonumber \\
    &=\mathbb{E}_{(u,v)}\left(\sum_{i,j}\phi(u_i)\mathbb{E}[\mathbf{A}_{ij}]\phi(v_j)\right) \nonumber \\
    &=\mathbb{E}_{(u,v)}\left(\sum_{i}\phi(u_i)\mathbb{E}[\mathbf{A}_{ii}]\phi(v_j)\right) \nonumber \\
    &=n_in_o\mathbb{E}_{(u,v)\sim\Omega(\rho)}\left(\phi(u_i)\phi(v_j)\right) \nonumber \\
    &=n_in_o\hat{\phi}(\rho)
\end{align}
With mean operation, we subtract average vector $w=\frac{1}{n_i}\sum_iW_i$ and have weight $\bar{\mathbf{W}}=\mathbf{W}-w\mathbf{1}^T$, we have $\bar{\mathbf{A}}_{ij}=(W_i-w)^T(W_j-w)=\mathbf{A}_{ij}-\Delta_{ij}$, where $\Delta_{ij}=w^T(W_i+W_j-w)$ with conditional expectation $\mathbb{E}[\Delta_{ij}|w]=w^Tw$. Recall $w$ is sum of Gaussian variable, as a result, $w\sim \mathcal{N}(0,1/n_i)$. And 
$\mathbb{E}[\Delta_{ij}|w]$ is a generalized Chi-Square distribution with expectation $\mathbb{E}_w[w^Tw]=\frac{n_o}{n_i} $. Then,
\begin{equation}
   \mathbb{E}_W[\Delta_{ij}]=\mathbb{E}_w\left(\mathbb{E}[\Delta_{ij}|w]\right)=\frac{n_o}{n_i},\quad \forall\ i,j
\end{equation}
With this result,

\begin{align}\label{eq.ap3}
    \mathbb{E}_{W}[\bar{\Phi}(\rho)]&=\mathbb{E}_{W}\mathbb{E}_{(u,v)}(\phi(u)^T\bar{\mathbf{A}}\phi(v)) \nonumber \\
    &=\mathbb{E}_{(u,v)}\left(\sum_{i,j}\phi(u_i)\mathbb{E}[\mathbf{A}_{ij}-\Delta_{ij}]\phi(v_j)\right) \nonumber \\
    &=\mathbb{E}_{W}[\Phi(\rho)]-\mathbb{E}_{(u,v)}\left(\sum_{i,j}\phi(u_i)\mathbb{E}[\Delta_{ij}]\phi(v_j)\right) \nonumber \\
    &=n_in_o\hat{\phi}(\rho)-\frac{n_o}{n_i}\mathbb{E}_{(u,v)}\left(\sum_{i,j}\phi(u_i)\phi(v_j)\right) \nonumber \\  
    &=n_in_o\hat{\phi}(\rho)-\frac{n_o}{n_i}\mathbb{E}_{(u,v)\sim \Omega(\rho)}(n_i\phi(u_i)\phi(v_i) \nonumber \\ & \quad +n_i(n_i-1)\phi(u_i)\phi(v_j)) \nonumber \\ 
    &=n_in_o\hat{\phi}(\rho)-\frac{n_o}{n_i}(n_i\phi(\rho)+n_i(n_i-1)\hat{\phi}(0)) \nonumber \\ 
    &=n_o(n_i-1)(\hat{\phi}(\rho)-\hat{\phi}(0))
\end{align}
Eq.~\ref{eq.ap2} and Eq.~\ref{eq.ap3} induce our assertion.


\section*{Appendix. B: Extended experiments}
\subsection*{B. 1: Straight deep network correlates input data}
\begin{figure}[h]
  \centering
  \includegraphics[width=0.35\textwidth]{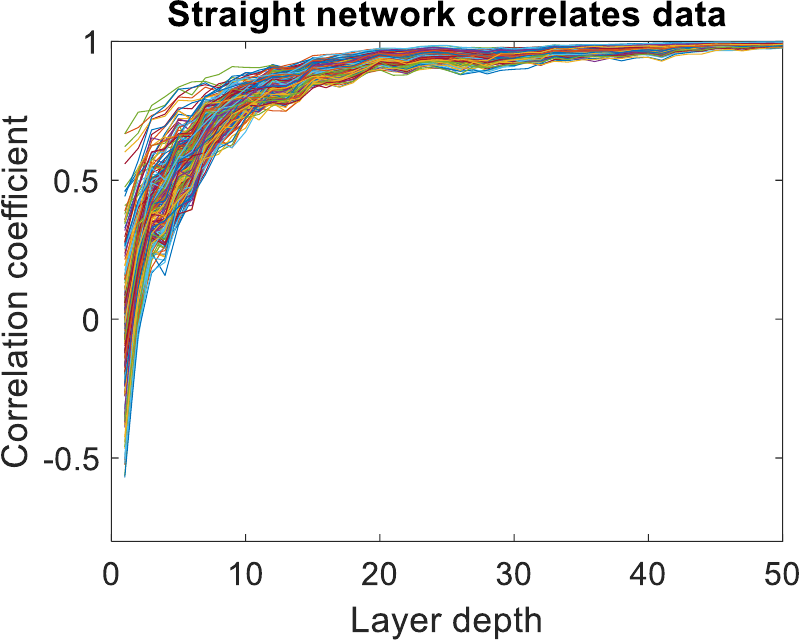}
  \caption{We calculate correlation coefficients on $200$ pairs of inputs for straight network. As the depth of layer increases, activations of all pairs tend to identical. }
  \label{fig.app1}
\end{figure} 
To explicitly show the freeze NNGP kernel, we calculate activations' correlation coefficients along layers. We use fully connected network to do handwritten digits recognition (MNIST dataset). Layer widths in our network are equally set to be $300$. We random sample $200$ pairs inputs from normalized training dataset and calculate their correlation coefficients on different layers. Fig.~\ref{fig.app1} shows the freeze phenomenon of NNGP kernel. 

\begin{figure}[ht]
  \centering
  \includegraphics[width=0.35\textwidth]{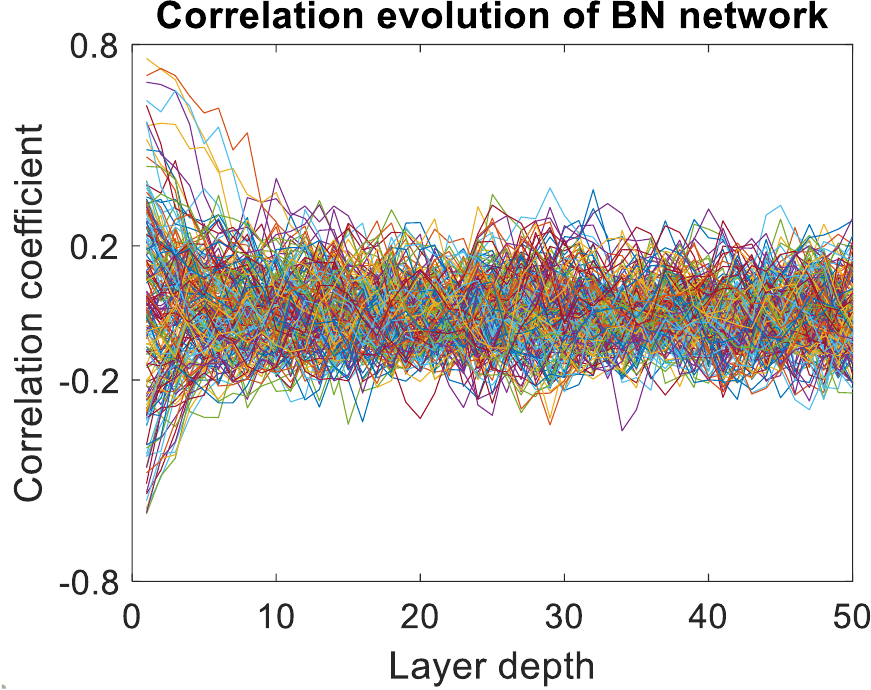}
  \includegraphics[width=0.35\textwidth]{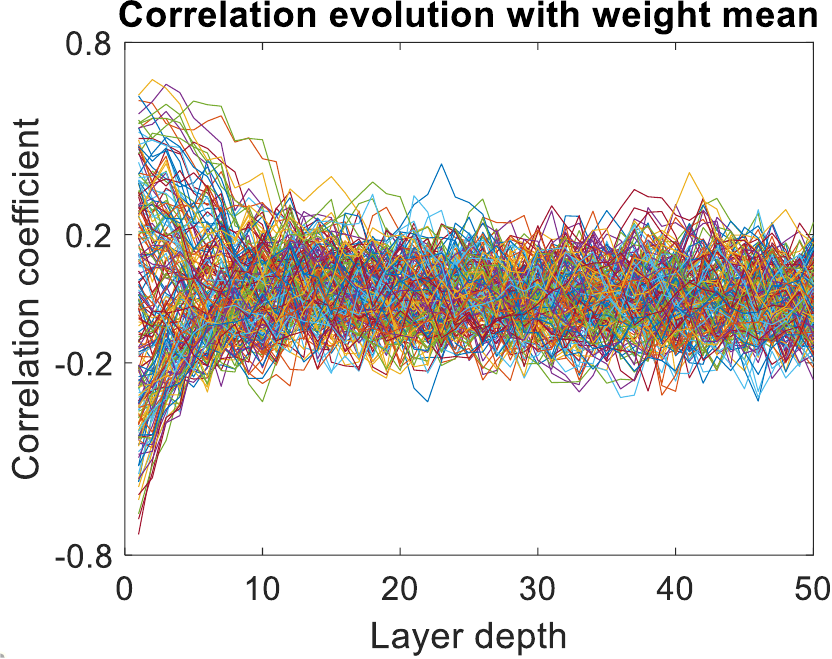}
  \caption{Networks with BN and weight mean show similar evolution as correlation coefficients concentrate at $[-0.2,0.2]$.}
  \label{fig.app2}
\end{figure}
Theoretic analysis by infinity width precisely describes the limited freeze NNGP kernel on straight networks. BN network and our method differ from the straight structure, and limit coefficients around zero. However, this empirical results show  a little discrepancy from theoretical prediction which states coefficients converge to $0$. Thus, infinite approximation like NTK may provide intuitive explanations and guidance for network design, the exact dynamic for realistic network is still an open problem. 

\subsection*{B. 2: Last BN layer stabilize variance}
\begin{figure}[ht]
  \centering
  \includegraphics[width=0.35\textwidth]{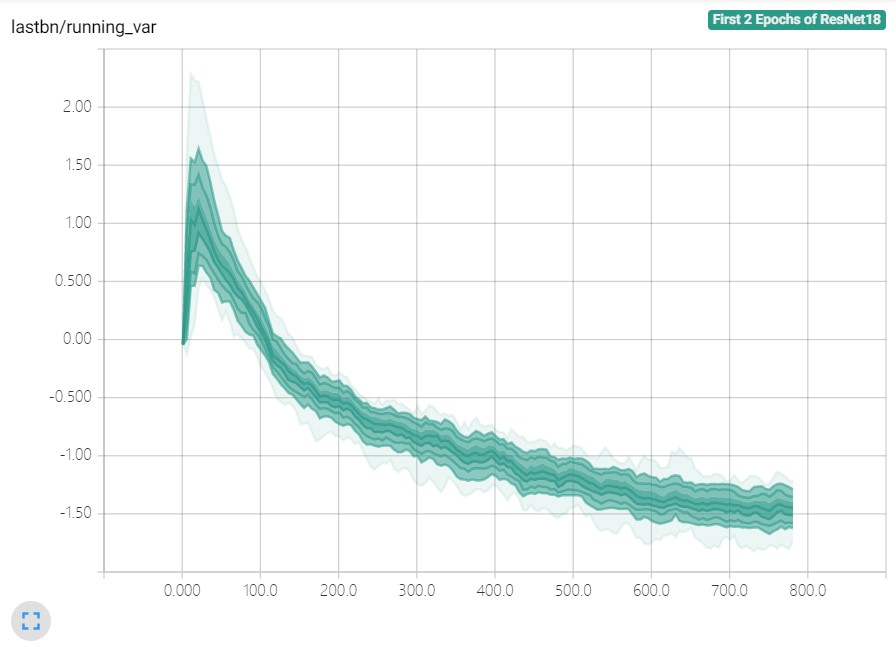}
  \includegraphics[width=0.35\textwidth]{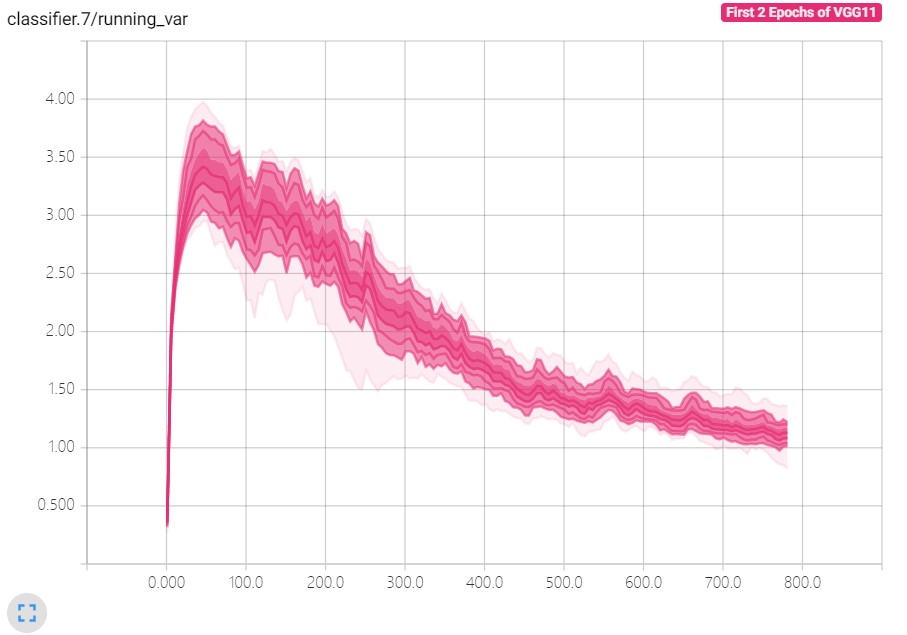}
  \caption{Running variance (logarithmic scales) of different channels in the output for VGG11 (left) and ResNet18 (right) networks. With larger learning rate, variance surges within several iterations. The last BN auto-tunes learning rate and variance becomes stable as network training. }
  \label{fig.app3}
\end{figure}
Straight network usually fail to train under large learning rates. We claim that one last BN layer contributes the trainability and we show the experimental evidence. We test on CIFAR100 dataset with VGG11 and ResNet18 structures using TensorBoard Package to investigate the running variance of last BN layer. We set learning rate as 0.1 and momentum 0.1 in our experiments, under which condition, straight structures diverge within several iteration. Fig.~\ref{fig.app3} records variance dynamic on first two epochs. Notice the y-axis is logarithmic scaled, large learning rate enlarges variance which causes the divergence of straight network, and last BN layer hinders this trend by auto-tuning learning rate.




\section*{Appendix. C: Further implementation optimization}
There are some potential optimizations to deploy our method. Firstly, mean weight operation doesn't need back propagation. As we all know, gradient of mean operation is also mean operation, such that, one execution either forward or backward mean operation is enough. Secondly, multiplier scalar at end of residual branch can be folded into convolutional layer, which would additionally reduce 5\%-10\% memory consumption. Typical Pytorch implementation of multiplying a learnable tensor stores buffers for input and product result. In-place multiplication would benefit memory utility. Last, as weight mean operation doesn't require float computation algorithm, full integral training is possible under our method, which may lead to $4\times$ thoughtput and bandwidth saving.

\end{document}